\documentclass[journal]{vgtc}                     

\onlineid{1174}

\vgtccategory{Area 4: Representations \& Interaction}

\title{Crystalis: Progressive Nucleation and Semantic Annealing for Coordinated Multi-View Visualization Generation}

\author{%
  Dazhen Deng, Zhaoping He, Xin Qian, Xiaotong Wang, Zi Ying, and Yingcai Wu
}

\authorfooter{
\item
    The authors are with Zhejiang University.
}

\abstract{%
  Large language models (LLMs) can generate individual charts, but coordinated multi-view visualizations (CMVs)---where views share data flows and cross-view interactions---remain out of reach: tight field-level coupling among data transformations, visual encodings, and interaction coordinations causes errors in one component to silently invalidate others. Rather than pursuing end-to-end analytical quality, which depends on model capability, domain knowledge, and user expertise, we target a foundational question: \textit{can LLMs reliably produce structurally correct CMVs, and what abstractions make this possible?} We present \system{}, a framework built on \textit{query-centric CMV modeling} that decomposes a CMV into structured queries over a dependency graph spanning three component types (Data, Visualization, Interaction) and three abstraction levels (requirement, specification, executable object). Two complementary mechanisms operate over this structure: \textit{progressive nucleation} crystallizes each query vertically from requirement to object along the dependency order, while \textit{semantic annealing} enforces horizontal consistency across queries at each level through layered logical checks. On a 12-task benchmark across five frontier LLMs, \system{} achieves up to 75\% end-to-end success---substantially outperforming an agentic coding baseline (8.3\% E2E with the same foundation model)---and a user study with 12 practitioners confirms the usability of the decomposition and iterative refinement workflow.

}


\teaser{
  \centering
  \includegraphics[width=\linewidth]{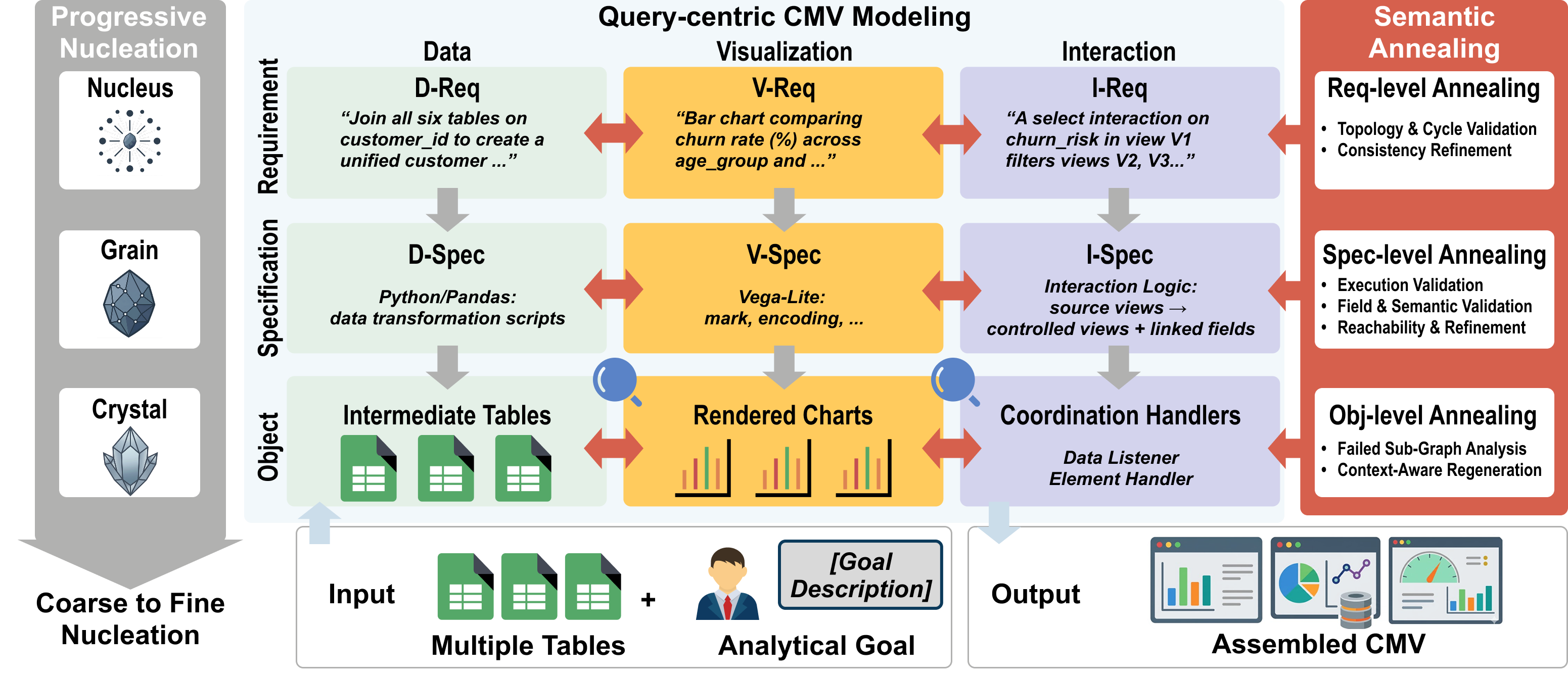}
  \caption{%
    \system{} models CMV generation as a matrix of structured queries across three query types (Data, Visualization, Interaction) and three abstraction levels (Requirement, Specification, Object). \textit{Progressive nucleation} (left) crystallizes each query vertically from compact requirements to executable objects, while \textit{semantic annealing} (right) enforces horizontal consistency across queries at each level through layered logical checks.
  }
  \label{fig:teaser}
}

\graphicspath{{figs/}{figures/}{pictures/}{images/}{./}}

\usepackage{tabu}
\usepackage{booktabs}
\usepackage{multirow}
\usepackage{lipsum}
\usepackage{mwe}
\usepackage{enumitem}
\usepackage{amsmath}
\usepackage{tabularx}
\usepackage{color}
\usepackage[normalem]{ulem}
\usepackage{algorithm}
\usepackage{algpseudocode}

\usepackage{mathptmx}

\newcommand{\system}{Crystalis}

\newcommand{\iconLogic}{\raisebox{-0.15em}{\includegraphics[height=0.95em]{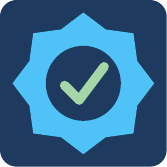}}\kern0.15em}
\newcommand{\iconLLM}{\raisebox{-0.15em}{\includegraphics[height=0.95em]{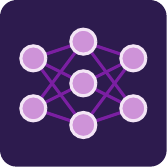}}\kern0.15em}

\begin{document}



\firstsection{Introduction}

\maketitle

\label{sec:intro}

Coordinated multi-view visualizations (CMVs) are a cornerstone of visual analytics~\cite{4269947,9222323}. By integrating multiple views through shared data flows and interactive coordination---where selecting in one view filters or highlights related records in others---CMVs enable analysts to examine complex datasets from complementary perspectives within a unified analytical context. Building effective CMVs, however, remains a demanding endeavor. Decades of research have produced rich theoretical foundations---the nested model~\cite{5290695}, design study methodologies~\cite{6327248}, and task taxonomies~\cite{6634168}---precisely because constructing these systems requires deep expertise spanning domain understanding, data preparation, chart design, and interaction orchestration~\cite{Keim2008,6875967}. As large language models (LLMs) demonstrate growing competence in generating individual visualizations~\cite{dibia2019data2vis,tian2024chartgpt,10753451}, a natural question arises: can LLMs also generate CMVs?

The answer is not straightforward. While LLMs can produce individual charts competently, a CMV is far more than a collection of independent views. Its views form a \textbf{coupled system}: each view draws from processed data whose schema determines what fields are available for encoding and interaction; cross-view coordinations such as linked brushing or filtering impose implicit contracts requiring that anchor fields exist in all participating views with consistent semantics; and upstream changes---an added aggregation, a renamed column---can silently invalidate downstream views and interactions elsewhere in the system. This \textbf{field-level coupling across components} is what distinguishes CMV generation from single-chart generation: each part's correctness depends on decisions made in other parts, and errors propagate silently across boundaries. Existing LLM-based visualization methods~\cite{deng2023dashbot,wu2021multivision,shi2020calliope,cheng2024snil} primarily produce loosely connected chart collections or narrative sequences where views do not share such tight structural dependencies, leaving the coupled CMV generation problem largely unaddressed.

Two natural strategies for applying LLMs to this problem each fall short in different ways. \textit{Direct code generation}---having an LLM (or an agentic coding assistant such as Claude Code or Cursor) produce a complete front-end/back-end implementation---buries cross-view dependencies in hundreds of lines of code, making failures hard to localize and field-level mismatches hard to detect. \textit{Platform-mediated generation}---having an LLM compose pre-built dashboard widgets through tool APIs (e.g., Tableau~\cite{981851})---constrains the output to fixed chart types and interaction templates, limiting the analytical flexibility that custom CMVs require. Neither provides an intermediate abstraction that is both compact enough for reliable LLM generation and expressive enough to capture the coupling that defines CMVs. We therefore ask: \textit{Can LLMs reliably generate structurally correct CMVs with functioning cross-view interactions, and what abstractions make this possible?}
The coupling gives rise to three concrete challenges.

\textbf{How to model CMV generation for LLMs.} A CMV specification can easily exceed hundreds of lines, and even a few erroneous parameters cause rendering failures. What is needed is a formulation that decomposes CMVs into tractable generation units---compact enough for LLMs to produce reliably---while preserving the inter-component dependencies that define their semantics.

\textbf{How to decouple generation while preserving dependencies.} Data transformations, visual encodings, and interaction coordinations are tightly coupled at the data-field level. Generating all components simultaneously leads to compounding errors, yet generating them independently risks breaking the contracts that bind them. The challenge is to decouple generation so that each component can be produced in isolation while respecting the dependency order that governs data flow.

\textbf{How to re-link decoupled components through logical validation.} Independently produced components may be individually correct yet globally inconsistent---a chart may reference a field that its upstream data processing dropped; an interaction may link views through fields with different semantic meanings. The challenge is to progressively verify that decoupled components satisfy the contracts that couple them.

We propose \textbf{\system{}}, which addresses these challenges through a modeling abstraction and a generation framework. To tackle the modeling challenge, we introduce \textbf{query-centric CMV modeling}, which abstracts CMV generation as the composition of structured queries over a dependency graph, where each query produces a component with an explicit schema contract that downstream queries can consume and validate against. 
Building on this model, the \textbf{\system{} framework} addresses the remaining two challenges through two complementary mechanisms. To decouple generation while preserving dependencies, \textbf{progressive nucleation}---analogous to how molecules in crystallography aggregate at seed sites to form ordered structures---refines each query vertically through three abstraction levels (requirements, specifications, executable objects) along the dependency order. To re-link decoupled components, \textbf{semantic annealing}---analogous to how thermal annealing relieves internal stresses in materials---primarily enforces horizontal consistency across queries at each abstraction level, detecting and correcting violations before they cascade, with a final cross-level recovery mechanism for persistent failures.

We implement \system{} as an interactive system where practitioners co-create CMV prototypes with LLMs, translating analytical requirements into data transformation code and Vega-Lite specifications with cross-view coordination~\cite{YE202443,ying2024vaid}. This work focuses on the \textit{modeling} and \textit{robust generation} of CMVs---ensuring that the framework can reliably produce structurally correct and semantically consistent outputs. The ultimate analytical quality of a generated CMV depends on factors beyond the generation framework itself---the foundation model's reasoning capability, domain knowledge available to the system, and the user's expertise in formulating goals and steering refinement. We view this work as an initial exploration that establishes the structural generation problem and a viable solution; our user study confirms that the D-V-I decomposition is intuitive and that practitioners can effectively steer generation through requirement editing. Our contributions are:
\begin{itemize}[leftmargin=*, label=$\diamond$]
\item \textbf{Query-centric CMV modeling} that abstracts CMV generation as the composition of structured queries over a dependency graph, decomposing a monolithic specification into compact, independently generable units with explicit schema contracts.

\item The \textbf{\system{} framework} with \textbf{progressive nucleation} and \textbf{semantic annealing}, which decouples CMV generation along the dependency order while progressively eliminating cross-component inconsistencies through interwoven logical checks.

\item \textbf{\system{} system}, an interactive CMV generation system that operationalizes the framework for human-centered visual analysis.

\item A \textbf{comprehensive evaluation} including a CMV generation benchmark, a comparative study across frontier LLMs, and an expert user study assessing usability and effectiveness.
\end{itemize}

\section{Related Work}
\subsection{Theoretical Models for Visual Analytics}
The knowledge generation model for visual analytics (VA)~\cite{Keim2008, 6875967} frames VA systems as artifacts that integrate models and visualization for knowledge discovery through iterative exploration. Munzner's nested model~\cite{5290695} decomposes visualization design into layered abstractions---domain characterization, data/operation abstraction, encoding/interaction design, and algorithm design---establishing a principled progression from high-level requirements to low-level implementation. Subsequent task typologies~\cite{6634168,6634156,1532136,8019880,8023762} and empirical studies~\cite{10874217} further bridge analytical goals with visualization techniques.
While these models guide \textit{human} designers, advancing \textit{generative} approaches for VA prototypes requires deeper structural modeling. VA systems are integrated codebases whose technical complexity makes direct LLM-based generation impractical, highlighting the need for intermediate abstractions that preserve system semantics while enabling automated generation. Ying et al.~\cite{ying2024vaid} explore this direction by building an index structure inspired by Vega-Lite to catalog VA view designs, expanding its property vocabulary to cover complex data transformations and composite encodings, demonstrating how structured annotations can characterize essential system components from data processing to coordinated views.

\subsection{Coordinated Multi-View Visualizations}
Coordinated multi-view visualizations (CMVs)~\cite{4269947} integrate multiple views through shared data flows and interaction logic, with most practical VA implementations adopting CMV architectures~\cite{9222323}. Prior research has established design guidelines~\cite{8017651,10.1145/345513.345271}, quantitative layout analysis~\cite{8933655,shao2021modeling}, and systematic composition patterns~\cite{9222323}.

A key challenge is \textit{view coordination}. North~\cite{north1997generalized} formalized coordination as inter-view linkages, leading to the Snap-together model~\cite{north2000snap} that coordinates views through primary-foreign key relationships~\cite{north2002schemas}. Recent work~\cite{10745882} extends this with graph-based representations modeling both interaction patterns and spatial adjacency.

Modern tools implement coordination through data-driven joins~\cite{north2000snap}, template-based systems such as Polaris~\cite{981851}, interaction-centric techniques~\cite{8017621,1382904}, and declarative grammars such as Vega-Lite~\cite{7539624}. Mosaic~\cite{heer2024mosaic} provides a cross-view coordination architecture that connects interactive views through a shared database layer, enabling scalable linked selections across heterogeneous visualizations. Observable Framework~\cite{observable2024framework} takes a complementary approach, providing a static site generator for data apps where data loaders, transformations, and interactive views compose into dashboards through reactive dataflow. While Mosaic and Observable focus on \textit{authoring infrastructure}---providing powerful primitives for developers to manually compose coordinated views---our work focuses on \textit{automated generation}: given an analytical goal, producing a structurally correct CMV via LLM-guided decomposition. We currently target Vega-Lite for its representativeness and LLM familiarity in single-view encoding; integrating richer coordination and dataflow frameworks such as Mosaic or Observable is a promising direction for extending output expressiveness.

\subsection{Visualization Recommendation and Generation}
Visualization recommendation systems suggest chart parameters for datasets through rule-based~\cite{mackinlay1987automatic}, browsing-based~\cite{wongsuphasawat2015voyager,wongsuphasawat2017voyager}, and learning-based~\cite{luo2018deepeye,hu2019vizml,dibia2019data2vis,9552844} approaches. NL-driven generation methods---from NL4DV~\cite{9222342} and NcNet~\cite{9617561} to LLM-based ChartGPT~\cite{tian2024chartgpt}---increasingly support visualization specification from textual descriptions~\cite{10.1145/3411764.3445400,10.1145/3472749.3474792,srinivasan2023bolt}. However, these methods primarily target single-view visualizations without cross-view coordination.

For multi-view generation, MultiVision~\cite{wu2021multivision} and DashBot~\cite{deng2023dashbot} incorporate hand-crafted design metrics within learning-based frameworks, while LightVA~\cite{10753451} integrates LLMs for task decomposition. MEDLEY~\cite{9911200} recommends dashboard collections based on analytical intents. However, cross-view coordination in these systems remains basic---mostly limited to Vega-Lite's built-in selection synchronization---and none models data transformation flows across views or interaction coordination flows between them. Nebula~\cite{9417674} constructs coordinated interactions from natural language but assumes predefined views; VisFlow~\cite{7536189} and FlowSense~\cite{8807265} support incremental dataflow construction but lack holistic coordination. Our work generates CMVs end-to-end: given multiple datasets and an analytical goal, it produces a CMV with explicit data transformation flows and interaction coordination flows.

\section{Preliminary Study}
\label{sec:preliminary}

To inform the design of \system{}, we conducted a preliminary study to understand how practitioners currently develop CMV systems and where LLM-based coding assistance falls short.

\paragraph{Participants and Protocol.}
We recruited six data analysis and visualization practitioners (P1--P6) via email, specifically targeting individuals with hands-on CMV development experience to obtain in-depth insights. Participants included two assistant professors (P1, P2), one postdoctoral researcher (P5), and three PhD students (P3, P4, P6), collectively with 2--8 years of VA experience. All had built multiple CMV systems and used LLM-based coding assistants daily. We conducted semi-structured interviews (30--50 minutes each) then performed thematic analysis~\cite{braun2006using} to identify recurring themes. The study was approved by the university's institutional review board.

\subsection{Questions and Findings}

\paragraph{Interview Questions.}
The interviews covered three areas: (1)~how participants decompose and organize a CMV project; (2)~where LLM-based coding assistants help and where they fall short; and (3)~common sources of cross-component bugs and rework. Three themes emerged.

\paragraph{F1: Staged workflows without formal intermediate representations.}
All participants described CMV development as a staged pipeline---data processing, visualization design, then interaction implementation---but diverged in how they orchestrate this with LLMs. P1 follows a \textit{generate-first} strategy: sketch a layout, have the LLM produce an initial implementation, then iterate by aligning weaker modules to well-working ones. P2 adopts the opposite \textit{plan-first} approach: specify the analytical goal, have the LLM plan each module, carefully review its reasoning, then authorize execution. P4 maintains a README as shared context: \textit{``I first have the LLM write a README\ldots so when making changes the LLM can take care of the whole repository.''} Others restrict LLMs to isolated tasks (P3) or break requirements into small units with explicit I/O specifications (P6). These strategies reveal a common gap: experts possess an implicit staged decomposition but lack a formal intermediate representation, leading to either monolithic prompts or ad hoc decomposition without structural contracts.

\paragraph{F2: Competent isolated generation, fragile cross-component consistency.}
LLMs perform well on bounded tasks but degrade across component boundaries. P5 trusts LLMs \textit{``mainly for simple visualization charts''} but not cross-component logic; P6 reported that LLMs \textit{``struggle with complex interaction logic involving frequent state changes.''} P4 identified the bottleneck: \textit{``The analysis flow is definitely not one-shot---you need to figure out what to look at first, then next. That process is the hardest.''} These observations point to a design where LLMs handle component-level generation while structured validation enforces cross-component consistency.

\paragraph{F3: Data flow and field-level consistency as core pain points.}
Participants consistently identified cross-component data flows as the most challenging aspect. P3: \textit{``The most complex part of coding is organizing the data flow between frontend and backend.''} P4 noted that \textit{``global consistency easily becomes mismatched---for example, changing the backend without updating the frontend.''} These field-level mismatches produce silent failures (P5: \textit{``backend errors are hidden and hard to debug''}), suggesting the core problem is the absence of explicit, validatable contracts at the data-field level.

\subsection{Design Considerations}

\begin{itemize}[leftmargin=*]
    \item \textbf{DC1: Abstract CMV structure for LLM generation (F1).}
    Formalize data, visualization, and interaction as compact, independently generable units with explicit inter-component dependencies.

    \item \textbf{DC2: Generate progressively from coarse to fine (F1, F2).}
    Refine components from requirements through specifications to executable objects, with validated outputs as context for subsequent stages.

    \item \textbf{DC3: Validate cross-component consistency at multiple levels (F2, F3).}
    Enforce contracts spanning semantic alignment, topology, and field-level compatibility through layered deterministic validation.

    \item \textbf{DC4: Support refinement with human oversight (F1, F2).}
    Allow practitioners to inspect and refine intermediate outputs at each stage, preserving the human judgment identified as irreplaceable.
\end{itemize}
\section{Query-Centric CMV Modeling}
\label{sec:modeling}

Following these considerations, we introduce \textit{query-centric CMV modeling}, which reformulates CMV generation as the composition of structured \textit{queries} over a \textit{dependency graph}. Three query types---data (D), encoding (V), and coordination (I)---each produce a component with an explicit \textit{schema contract} that downstream queries consume. These contracts transform the implicit field-level couplings identified in our preliminary study into checkable constraints at query boundaries.

\subsection{Components and Field-Level Coupling}
\label{subsec:coupling}
\paragraph{Input and Output}
Given structured tables and a textual analytical goal as input, the framework generates a functional CMV comprising three types of components: \textbf{data transformations (D)} that map input tables to intermediate tables via Python/pandas; \textbf{visual encodings (V)} that map data attributes to visual channels via Vega-Lite; and \textbf{interaction coordinations (I)} that define how user actions in one view propagate to others.

These components are \textbf{tightly coupled at the data-field level} in three ways.
\textit{Field availability} across D$\rightarrow$V: aggregations silently drop fields from intermediate tables, causing downstream views to reference nonexistent columns.
\textit{Field consistency} across V$\rightarrow$I: interaction coordinations link views through shared anchor fields that must exist with consistent semantics across all participating views.
\textit{Implicit field propagation} across D$\rightarrow$V$\rightarrow$I: interaction requirements impose constraints on upstream data transformations---e.g., a cross-view filter requires the anchor field to be preserved even if the view does not encode it---creating dependencies that flow \textit{against} the generation order.

\paragraph{Dependency Graph}
\label{subsec:dependency_graph}
We model the CMV as a directed dependency graph $G = (N, E)$, where $N = N_D \cup N_V \cup N_I$ and edges follow a strict layered topology: each $v \in N_V$ connects to exactly one $d \in N_D$ (D$\rightarrow$V), and each $i \in N_I$ connects to its participating views (V$\rightarrow$I). The view-level graph induced by interactions must be acyclic. No D$\rightarrow$D, V$\rightarrow$V, or I$\rightarrow$I edges are permitted. This graph serves as the backbone for both generation (following topological order) and validation (checking consistency over edges) in the framework (\autoref{sec:framework}).

\subsection{Query Abstraction and Schema Contracts}
\label{subsec:query_model}

Prior work models visualization execution as dataflows~\cite{7536189}, declarative queries~\cite{9552893}, or function composition~\cite{wu2024designspecifictransformationsvisualization}. We apply the same compositional perspective to CMV \textit{generation}: rather than monolithic code synthesis, we model it as composing structured queries over the dependency graph, where each query consumes upstream schemas and produces a component with an explicit output contract (\textbf{DC1}).

At the \textbf{object level}, the three query types correspond to the three node types in $G$:
\begin{itemize}[leftmargin=*]
    \item \textbf{Data query} ($d \in N_D$): A transformation producing an intermediate table with an explicit schema (column names and types).
    \item \textbf{Encoding query} ($v \in N_V$): A mapping from one intermediate table's fields to visual channels, bound to its upstream data query's schema.
    \item \textbf{Coordination query} ($i \in N_I$): A cross-view coordination declaring source views (with selection type and link field), controlled views (with target field and action type), and the shared anchor fields through which selections propagate.
\end{itemize}

At the \textbf{field level}, schema contracts formalize the couplings from \autoref{subsec:coupling}: data queries publish output schemas; encoding queries declare which fields they bind; coordination queries specify anchor fields that must exist with consistent semantics across all participating views. This makes field-level dependencies a first-class, validatable part of the model.
Each query is compact (tens of lines rather than hundreds), independently validatable against its neighbors, and composed following the dependency order D$\rightarrow$V$\rightarrow$I. How to progressively refine each query and enforce cross-query consistency is the subject of the \system{} framework (\autoref{sec:framework}).
\section{\system{} Framework}
\label{sec:framework}

Building on the query-centric model, we present the \system{} framework that refines each query through three \textbf{abstraction levels}---\textit{requirement} (natural-language description), \textit{specification} (structured schema: D-Spec, V-Spec, or I-Spec), and \textit{executable object} (intermediate table, Vega-Lite fragment, or coordination code)---yielding a $3 \times 3$ matrix across query types and abstraction levels (\textbf{DC2}).

At each cell, two complementary mechanisms interleave, drawing on a crystallographic metaphor. \textbf{Progressive nucleation}---analogous to molecules aggregating at seed sites to form an ordered crystal---crystallizes each query from requirement through specification to executable object, following the dependency order D$\rightarrow$V$\rightarrow$I and Munzner's nested model~\cite{5290695}. \textbf{Semantic annealing}---analogous to controlled heat eliminating lattice defects---enforces cross-query consistency through layered validation checks that tighten from structural integrity to field-lineage correctness to semantic consistency (\textbf{DC3}).

As illustrated in \autoref{fig:teaser}, the three abstraction levels correspond to successive stages of crystal growth: a \textit{nucleus} (compact requirement seed), a \textit{grain} (structured specification with defined facets), and a \textit{crystal} (fully formed executable object). The matrix is traversed column by column: nucleation generates, annealing validates, and errors are fed back before advancing---forming a tight generate-validate loop at each cell. Throughout the following subsections, we mark each check as either \iconLogic\textit{programmatic} (deterministic logic) or \iconLLM\textit{LLM-based} (semantic reasoning).

\subsection{Requirement Level: Nucleus Generation}
\label{subsec:req_level}

The requirement level forms the \textit{nucleus}---a compact seed that encodes the analytical intent. It translates the user's goal into a structured dependency graph of natural-language descriptions, one per query.

\subsubsection{Requirement-Level Nucleation}
Rather than generating all requirements simultaneously, nucleation at this level proceeds in two phases:

\textbf{Data and Visualization Requirements (D+V).}
Given the input tables and the analytical goal, the LLM generates a set of \textit{data requirements} and \textit{visualization requirements} along with their dependencies. Each data requirement specifies a data transformation (e.g., aggregation, filtering, joining) that produces an intermediate table, while each visualization requirement specifies a chart type and visual encoding for a particular analytical insight. Each V connects to exactly one D (one intermediate table), forming a bipartite D$\rightarrow$V dependency structure.

\textbf{Interaction Requirements (I).}
Unlike D+V requirements, interaction requirements are generated \textit{incrementally}---one at a time---so that the interaction topology can be validated after each addition.
LLMs struggle to produce graph structures satisfying global constraints in a single pass. Interactions induce a view-level dependency graph (V$\rightarrow$V edges), which must remain acyclic: a cycle (e.g., V1 filters V2, V2 highlights V3, V3 filters V1) would create infinite triggering loops at runtime. Enforcing acyclicity is a global constraint that LLMs cannot reliably satisfy when generating all interactions at once. By generating one at a time and validating after each addition, the framework rejects cycle-introducing edges immediately and guides the LLM to propose alternatives.
Each interaction requirement specifies a source view (where user interaction occurs) and one or more target views (where the interaction effect propagates), forming V$\rightarrow$I edges in the dependency graph.

\subsubsection{Requirement-Level Annealing}
\begin{figure}[tb]
  \centering
  \includegraphics[width=\linewidth]{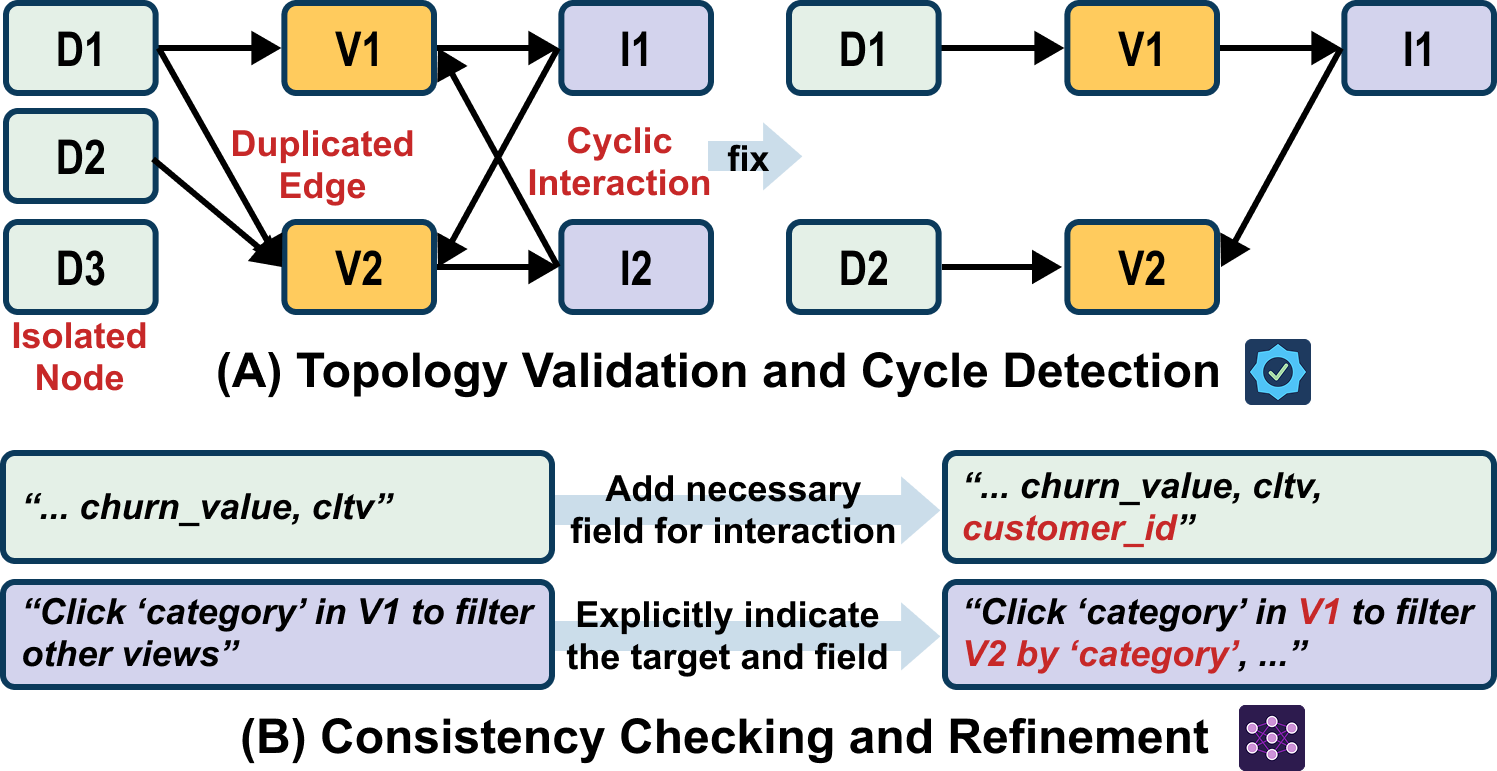}
  \caption{Requirement-level annealing. (A) Topology validation removes isolated nodes and duplicated edges, and cycle detection eliminates cyclic interactions to ensure a valid DAG structure. (B) Consistency refinement aligns field references across requirements---e.g., adding necessary fields to data requirements for downstream interactions and making interaction targets explicit.}
  \label{fig:req-annealing}
\end{figure}

After generation, annealing validates the requirement dependency graph through two checks (\autoref{fig:req-annealing}).

\textbf{Topology Validation and Cycle Detection~\iconLogic.}
After D+V requirements are generated, the framework verifies the dependency graph structure: only D$\rightarrow$V and V$\rightarrow$I edge patterns are allowed; every requirement must participate with at least one edge; each V connects to exactly one D, and each D has at least one downstream V. Text mentions are cross-checked against edges---if a requirement references another by name, the corresponding edge must exist. When violations are detected (e.g., isolated nodes or duplicated edges as in \autoref{fig:req-annealing}A), the error report is fed back to the LLM for correction.
After each interaction requirement is generated incrementally, we construct a view-level dependency graph (V$\rightarrow$V edges derived from interactions) and detect cycles using depth-first search, as well as transitive redundancy. A set of \textit{forbidden edges} is maintained to prevent the LLM from regenerating previously rejected interactions.

\textbf{Consistency Checking and Refinement~\iconLLM.}
After all requirements pass structural validation, the LLM reviews them jointly to enforce cross-query consistency (\autoref{fig:req-annealing}B). This includes: aligning field naming conventions across D, V, and I requirements; clarifying each view's role in interactions (source vs.\ target); and ensuring that fields required for cross-view coordination are explicitly mentioned in the relevant data requirements rather than left implicit in the visualization or interaction layer.

\subsection{Specification Level: Grain Generation}
\label{subsec:spec_level}

The specification level grows each nucleus into a \textit{grain}---a structured, validatable schema with defined facets that can be generated, validated, and assembled independently. This decoupling separates concerns that a monolithic Vega-Lite specification entangles: data loading, visual encoding, and interaction logic. Specifications are generated and immediately executed in D$\rightarrow$V$\rightarrow$I dependency order, so that each query's upstream outputs are available as context for the next. Within each type, queries are generated independently (e.g., all D-Specs are produced separately, each with its own dependency-aware context).

\subsubsection{Specification-Level Nucleation}

Each query type has a corresponding specification format that serves as its structured, validatable representation.

\textbf{Data Specification (D-Spec).}
Each D-Spec is an executable Python/pandas script that transforms input tables into a single intermediate table; Python is chosen for its extensibility to scientific computing libraries (e.g., scikit-learn for dimensionality reduction). The LLM generates the script, which is immediately executed; the output is captured as a CSV file along with its schema (column names and types). A D-Spec exposes a \texttt{table\_info} interface containing the output table name and its column list, which serves as the field-level contract that all downstream queries must respect.

\textbf{Visualization Specification (V-Spec).}
Each V-Spec is a \textit{partial} Vega-Lite specification that describes a single view's visual encoding without interaction logic. It contains a \texttt{markType} (e.g., \texttt{bar}, \texttt{point}, \texttt{arc}), a \texttt{table} reference to the intermediate table from its upstream D-Spec, and an \texttt{encoding} object mapping data fields to visual channels (\texttt{x}, \texttt{y}, \texttt{color}, \texttt{size}, etc.). Critically, a V-Spec \textit{excludes} all interaction-related constructs---no \texttt{params} for selections, no parameter-driven \texttt{transform} filters, no conditional \texttt{opacity}---ensuring interaction logic is generated with full knowledge of all participating views rather than prematurely embedded in individual view specs. We adopt Vega-Lite for its representativeness in single-view encoding and broad LLM familiarity; the D-V-I decomposition is grammar-agnostic, and richer coordination grammars such as Mosaic~\cite{heer2024mosaic} could replace or augment the current V-Spec and I-Spec representations.

\textbf{Interaction Specification (I-Spec).}
Each I-Spec defines a cross-view coordination as a mapping from \textit{source views} to \textit{controlled views}. For each source view, the spec declares the selection type (\texttt{point} or \texttt{interval}), the monitored encoding channels, and the \texttt{link\_field}---the data field whose selected value drives the coordination. For each controlled view, it declares the target \texttt{field} and the \texttt{action} (\texttt{filter} or \texttt{highlight}). This makes the cross-view data contract explicit: the \texttt{link\_field} and the target \texttt{field} must exist in their respective intermediate tables and share the same semantic category---a contract verified during specification-level annealing.

\textbf{Dependency-Aware Context.}
To proactively reduce field-level coupling failures, each specification is generated with a context drawn from its neighbors in the dependency graph. For a D-Spec, the context includes its downstream V and I requirements, along with \textit{shared fields}---fields extracted from interaction descriptions that must appear in the intermediate table for cross-view coordination to function (e.g., a \texttt{category} field needed by a downstream filtering interaction). For a V-Spec, the context includes its upstream \texttt{table\_info} and its interaction role (source, target, or both). For an I-Spec, the context includes the encoding and column lists of all participating views. This proactive mechanism complements annealing's reactive checks: dependency-aware context reduces the likelihood of errors, while validation catches those that still occur.

\subsubsection{Specification-Level Annealing}
\begin{figure}[tb]
  \centering
  \includegraphics[width=\linewidth]{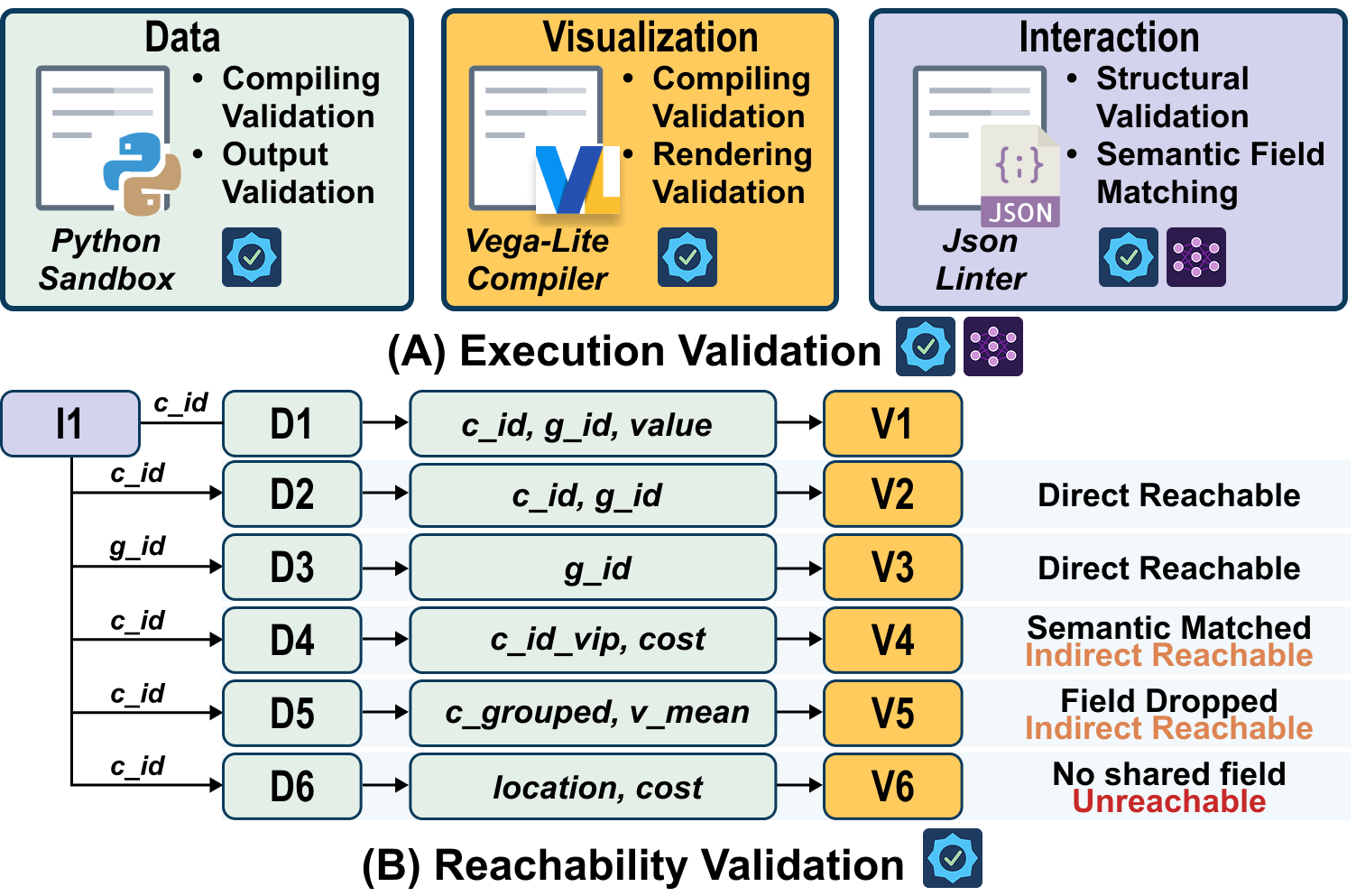}
  \caption{Specification-level annealing. (A)~\iconLogic Execution validation catches D-Spec runtime errors and V-Spec rendering failures, retrying with error feedback. (B)~\iconLogic Field consistency validation checks field existence, verifies structural alignment with the requirement-level topology, enforces semantic category matching, and validates data flow paths between source and target intermediate tables.}
  \label{fig:spec-annealing}
\end{figure}

After each specification is generated and executed, annealing validates it for executability and cross-query data-flow consistency, retrying with error feedback when violations are detected (\autoref{fig:spec-annealing}). The retry budget is set to 6 attempts for D-Specs (which involve longer code and more runtime failure modes) and 3 attempts each for V-Specs and I-Specs; requirement-level interaction generation also allows up to 3 attempts per interaction.

\textbf{Execution Validation~\iconLogic\iconLLM and Regeneration~\iconLLM.}
D-Specs are immediately executed after generation: the Python/pandas code is run against the input tables, and runtime errors (missing columns, type mismatches, malformed joins) trigger retry with the error traceback fed back to the LLM. V-Specs undergo test rendering via the Vega-Lite runtime to catch syntax errors, invalid mark types, or encoding mismatches before downstream queries consume them.

Unlike D-Specs and V-Specs, which benefit from established runtimes (Python and Vega-Lite) that catch errors automatically, I-Specs have no standard compiler or renderer. The framework therefore constructs a custom validation pipeline for I-Specs:
\begin{itemize}[leftmargin=*]
    \item \textit{Structural validation~\iconLogic}: the source and target views declared in the I-Spec must match those in the requirement-level dependency graph; any mismatch (extra, missing, or duplicated views) is flagged.
    \item \textit{Semantic category matching~\iconLLM}: after all D-Specs have been executed, the LLM classifies each column across all intermediate tables into a semantic category (e.g., \texttt{customer\_id} $\mapsto$ ``customer identifier'', \texttt{month} $\mapsto$ ``temporal''), identifying columns that represent the same real-world unit across different tables. Each I-Spec's \texttt{link\_field} and target \texttt{field} are looked up in this mapping---a missing entry indicates the field does not exist in the intermediate table, while a category mismatch (e.g., linking a temporal field to a categorical identifier) indicates a cross-view coupling error. All source views' \texttt{link\_field} values must share the same semantic category, and each target view's \texttt{field} must match that category.
\end{itemize}
Validation errors are fed back to the LLM for I-Spec regeneration, following the same retry loop as D-Specs and V-Specs.

\textbf{Reachability Validation~\iconLogic and Cross-Layer Refinement~\iconLLM.}
For each source--target field pair in an I-Spec, the framework performs a reachability query on the schema graph---a graph whose nodes are the input tables and whose edges are foreign-key relationships. A path is \textit{direct} if the target intermediate table already contains the anchor field, or \textit{indirect} if the two intermediate tables trace back to input tables connected through foreign keys. When an indirect path is found, the framework triggers cross-layer refinement: it augments the responsible D-Spec's requirement to include the missing join key, re-executes the D-Spec to produce an updated intermediate table, and re-validates the I-Spec against the new schema---resolving the implicit D$\rightarrow$V$\rightarrow$I field propagation problem identified in \autoref{subsec:coupling}. This repair modifies D-Specs and I-Specs but leaves V-Specs unchanged, since visual encodings do not depend on interaction anchor fields. When no path exists, the coordination is flagged as infeasible.

\subsection{Object Level: Crystal Composition}
\label{subsec:obj_level}

The object level fuses validated grains into a complete \textit{crystal}---a renderable CMV.

\subsubsection{Object-Level Nucleation}
Unlike the requirement and specification levels where the LLM generates content, object-level nucleation is purely programmatic.

\textbf{Interaction Injection~\iconLogic.}
For each I-Spec, the framework programmatically modifies the partial V-Specs to embed interaction logic: source views receive selection parameters, while target views are restructured to respond to those selections---filtering the displayed data or adjusting visual emphasis according to the specified action type.

\textbf{Coordination Code Generation~\iconLogic.}
For each interaction, the framework generates JavaScript coordination code that preloads the intermediate tables, binds signal listeners to each source view's selection parameters, computes the intersection of selected values across multiple sources when needed, and dynamically updates target views with filtered data. The assembled Vega-Lite specifications and JavaScript code together form the renderable CMV.

\subsubsection{Object-Level Annealing}
\begin{figure}[tb]
\centering
\includegraphics[width=\columnwidth]{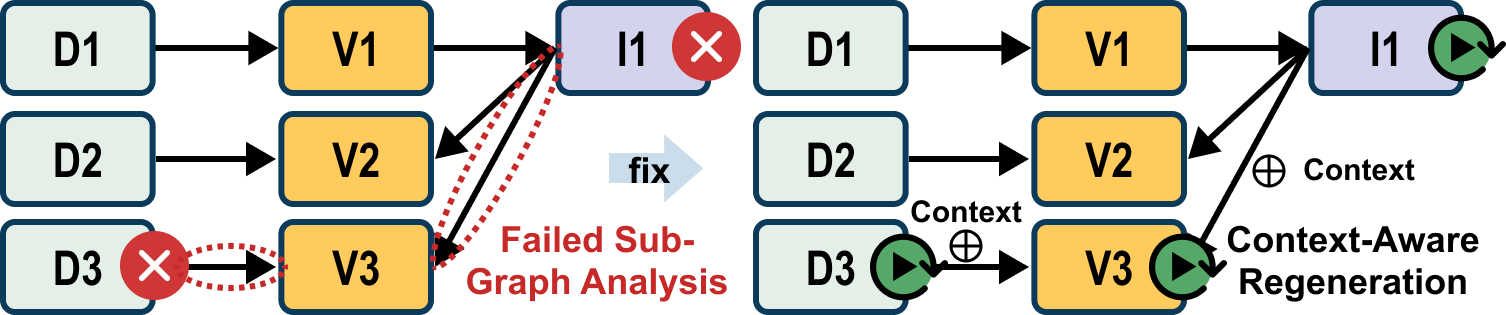}
\caption{Object-level annealing. \textit{Left}~\iconLogic: failed sub-graph analysis identifies failed nodes (D3, I1) and traces downstream via BFS to determine the affected sub-graph (red dashed arrows), stopping at interaction nodes to prevent over-expansion. \textit{Right}~\iconLLM: context-aware regeneration rewrites requirement descriptions using neighbor context and re-executes affected queries in dependency order (D$\rightarrow$V$\rightarrow$I).}
\label{fig:obj-annealing}
\end{figure}

Some queries may still fail after exhausting specification-level retries---a D-Spec that cannot produce a valid intermediate table, a V-Spec that fails to render, or an I-Spec whose fields remain inconsistent. These failures often stem from the requirement descriptions themselves rather than code generation errors. As illustrated in \autoref{fig:obj-annealing}, the regeneration mechanism addresses them through two steps.

\textbf{Failed Sub-Graph Analysis~\iconLogic.}
The framework identifies all D, V, and I queries that exhausted local retries, then traces downstream from each failed node via breadth-first search to determine the affected sub-graph. Critically, propagation \textit{stops at interaction nodes} to prevent over-expansion: if D2 and V2 fail, the downstream I1 (which depends on V2) is marked as affected, but I1's other participants (e.g., D1, V1) are \textit{not} pulled into the affected set, even though they participate in the same interaction. This bounding ensures that a localized failure does not cascade into a full pipeline rebuild.

\textbf{Context-Aware Regeneration~\iconLLM.}
For each affected node, the framework builds a \textit{neighbor context} that includes the requirements, execution status, error messages, and output schemas of its upstream and downstream neighbors. The LLM uses this context to rewrite the requirement description, addressing the root cause of failure while maintaining compatibility with successfully executed neighbors. The affected queries are then re-executed in dependency order (D$\rightarrow$V$\rightarrow$I) using the rewritten requirements.

\section{\system{} Interface and Case Study}
\label{sec:case}

\begin{figure*}[tb]
\centering
\includegraphics[width=\textwidth]{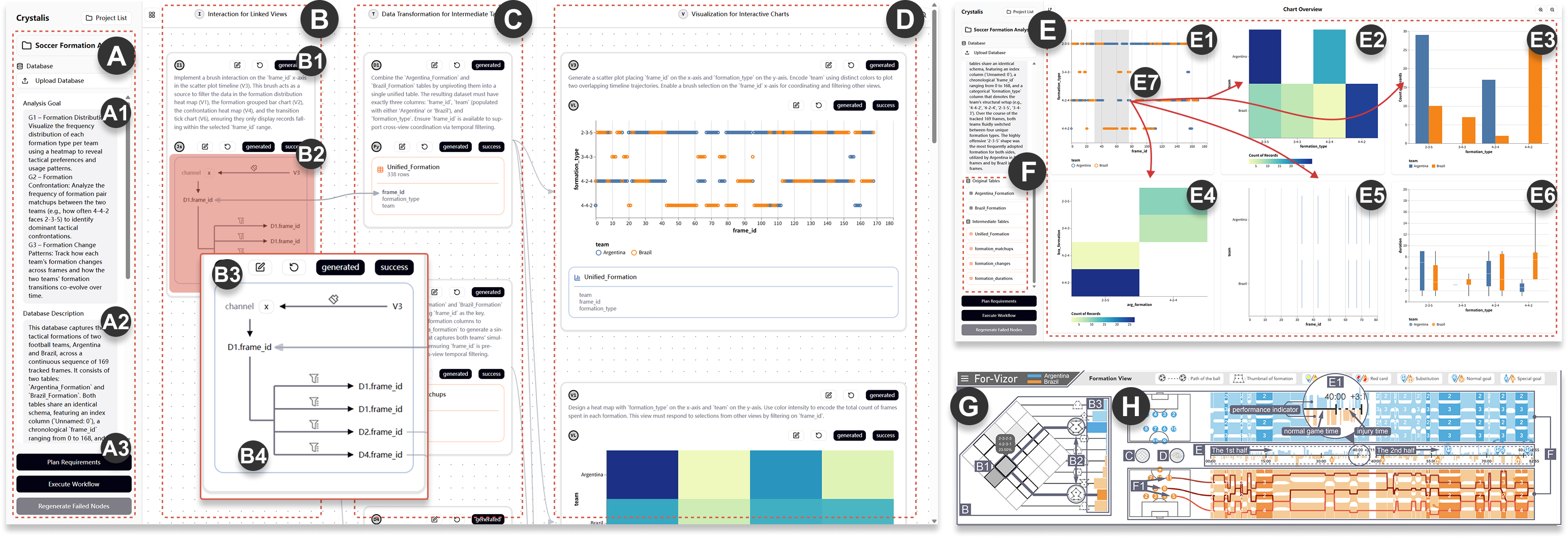}
\caption{\system{} interface and case study reproducing ForVizor~\cite{8440804}. (A)~Configuration panel. (B--D)~Query panels for interactions, data transformations, and visual encodings. (E)~Generated CMV dashboard. (F)~Source and intermediate tables. (G,~H)~Original ForVizor views for comparison.}
\label{fig:case}
\end{figure*}

\system{} provides a web-based interface (\autoref{fig:case}) that supports the full CMV generation workflow with human oversight (\textbf{DC4}). 

The configuration panel lets users specify an analytical goal~(A1) and input datasets~(A2), then trigger the three-stage pipeline---requirement generation, specification generation, and object refinement---via dedicated controls~(A3). The remaining panels expose intermediate outputs for each query type. Taking the interaction panel~(B) as an example: the requirement description~(B1) is editable---users can revise it and regenerate downstream code; the generated specification code~(B2) is also directly editable and re-executable. A field-level dependency graph~(B3) visualizes how views and intermediate tables are linked through interaction triggers, helping users inspect coordination details~(B4). The data transformation panel~(C) and visual encoding panel~(D) follow the same two-tier structure of editable requirement and code, enabling fine-grained intervention at any abstraction level. The generated CMV~(E) is rendered as an interactive dashboard, and a data table panel~(F) provides access to both source tables and generated intermediate tables.

To demonstrate end-to-end capability, we use \system{} to partially reproduce ForVizor~\cite{8440804}, a visual analytics system for exploring spatio-temporal team formations in soccer. We use ForVizor's formation-by-timestamps data, consisting of two CSV tables recording the formations of Argentina and Brazil, respectively. Given an analytical goal describing formation comparison and temporal evolution, \system{} generates a six-view CMV~(E1--E6). Views E2--E4 present formation confrontation statistics, serving a similar overview function to ForVizor's heatmap with bar charts~(G). Views E1, E5, and E6 approximate temporal formation dynamics through distribution charts and change-point detection, corresponding to ForVizor's formation change timeline~(H). The views are coordinated~(E7): selecting a time interval in the E1 timeline highlights the corresponding data across other views, enabling cross-view temporal exploration. We omit ForVizor's player-detail views, pitch video overlays, and dynamic trajectory animations, as these involve system-level operations beyond the scope of CMV generation. Compared to ForVizor's novel, purpose-built visual designs, the generated views naturally lack some visual expressiveness; however, this case illustrates how \system{} can serve as a rapid prototyping tool that helps practitioners bootstrap analytical CMV systems before investing in bespoke visual designs.

\section{Evaluation}
\label{sec:evaluation}

We evaluate \system{} through three complementary studies. A \textbf{CMV generation benchmark} (\autoref{subsec:benchmark}) tests the framework across multiple frontier LLMs and quantifies each annealing layer's contribution through generation-process instrumentation. A \textbf{baseline comparison} (\autoref{subsec:baseline}) contrasts \system{} with a strong agentic coding assistant to isolate the benefit of structured generation. A \textbf{user study} (\autoref{subsec:user_study}) assesses system usability through interactive sessions with data analysis and visualization practitioners.

\subsection{CMV Generation Benchmark}
\label{subsec:benchmark}

\subsubsection{Benchmark Construction}
We assembled a benchmark of 12 CMV generation tasks spanning diverse analytical domains, with datasets sourced from Kaggle and VA research publications. Each task consists of 2--6 input CSV tables and a natural-language analytical goal. The datasets cover sports analytics, e-commerce, healthcare, telecommunications, geography, entertainment, and finance, providing diversity in data schema complexity, table sizes, and the analytical operations required (aggregation, joining, filtering, dimensionality reduction). Task complexity varies in the number of input tables, the number of D/V/I queries generated, and the interaction patterns required (filtering, highlighting, brushing).


\subsubsection{Multi-LLM Comparison}

We evaluate the framework with five frontier LLMs---Claude Sonnet 4.6, GPT-5.4, Gemini 3.1 Pro, DeepSeek-V3.2, and Qwen-Plus---using the same benchmark tasks, prompts, and retry limits. A traditional ablation study---disabling one annealing layer at a time---is infeasible because removing early-stage annealing causes cascading failures that prevent downstream generation from executing at all. Instead, we instrument the generation pipeline to record the number of errors introduced by the LLM and resolved by annealing at each abstraction level, providing a fine-grained view of each layer's error-correction behavior \textit{in situ}. \autoref{tab:benchmark_results} reports both end-to-end performance and per-level annealing effectiveness.

\subsubsection{Generation Performance}

The \textbf{E2E} column reports the percentage of tasks producing a complete, renderable CMV with functioning interactions. Claude Sonnet 4.6 achieves the highest E2E success rate (75.0\%), followed by Gemini 3.1 Pro (50.0\%), GPT-5.4 (41.7\%), Qwen-Plus (33.3\%), and DeepSeek-V3.2 (25.0\%). E2E success requires \textit{all} D, V, and I components to execute correctly and coordinate without errors---a strict criterion that amplifies even small per-component failure rates, explaining the wide gap between per-component and E2E metrics.

The \textbf{D}, \textbf{V}, and \textbf{I} columns report per-component success rates in the format \textit{first-pass\,/\,post-spec-annealing\,/\,post-obj-regeneration}.
All models achieve high D success rates after annealing (75.9--100\%), though first-pass rates vary widely (49.2--83.9\%), indicating that data transformation code frequently requires retry---particularly for GPT-5.4 (49.2\% first-pass), which introduces the most D-Spec runtime errors.
V success rates are generally high (74.1--96.2\% post-spec) and show minimal improvement from spec to obj annealing (at most 5.2pp), confirming that V-Spec failures are largely resolved during specification-level retry.
I success rates are consistently the lowest (35.0--77.8\% post-spec), confirming that interaction specification---which requires reasoning about cross-view field coupling and coordination semantics---is the most challenging query type. Object-level regeneration provides meaningful recovery for I components, particularly for DeepSeek (+10.0pp) and Qwen (+12.5pp).

The \textbf{Req.}\ and \textbf{Spec.}\ columns report \textit{before/after} annealing success rates. At the requirement level, all models reach near-100\% after annealing, indicating that topology validation and consistency refinement effectively resolve structural errors. At the specification level, success rates improve by 7--14pp after annealing (e.g., DeepSeek: 57.4$\to$69.1\%), though the gap to perfect resolution reflects fundamentally difficult errors that exhaust the local retry budget. The \textbf{Obj.}\ column reports the final node success rate after object-level regeneration, ranging from 75.0\% (DeepSeek) to 95.3\% (Claude).

Token consumption does not correlate with performance: GPT-5.4 uses 30\% more tokens (363K) than Claude (280K) yet achieves a substantially lower E2E rate, suggesting that token efficiency depends more on first-pass quality than on model verbosity.

\subsubsection{Annealing Effectiveness}

The annealing columns report errors resolved out of those introduced (Res./Int.) and the resolution rate (RR.\%) at each level, quantifying each layer's error-correction contribution \textit{in situ}.

\textbf{Requirement-level annealing} catches topology violations (orphan nodes, invalid edges, duplicates) and cycle-introducing interactions, achieving near-perfect resolution rates (94.6--100\%). All models introduce substantial structural errors (28--48 per run), yet nearly all are corrected after error feedback---confirming that LLMs can reliably fix graph constraints given explicit error descriptions.

\textbf{Specification-level annealing} catches D-Spec runtime errors, V-Spec rendering failures, and I-Spec validation failures (field-existence violations, semantic category mismatches, structural mismatches), showing moderate resolution rates (37.2--59.1\%). Claude achieves the highest rate (59.1\%, 13/22), while DeepSeek has the lowest (37.2\%, 16/43). The lower rates reflect that specification-level failures often involve incorrect data transformations or field-coupling mismatches requiring substantial code changes.

\textbf{Object-level annealing} targets queries that have exhausted specification-level retries, with resolution rates of 0--32.0\%. DeepSeek (32.0\%, 8/25) and Claude (20.0\%, 2/10) benefit the most; Gemini achieves 0\%, suggesting its failures stem from capability limitations rather than requirement-level ambiguity. The overall RR.\% ranges from 56.2\% (DeepSeek) to 76.4\% (Claude).

Two cross-model patterns emerge. First, the number of specification-level errors correlates inversely with E2E success: Claude (22) and Gemini (23) introduce fewer errors than GPT (44) and DeepSeek (43), indicating that higher first-pass generation quality directly reduces the burden on downstream annealing. Second, models with fewer errors also achieve higher resolution rates, creating a compounding advantage---fewer errors \textit{and} better recovery---that explains the wide gap in final Obj.\ success rates (75.0--95.3\%). Notably, Gemini's V-Spec success rate remains flat across all three stages (78.7\%), and its object-level resolution rate is 0\%, suggesting that its remaining failures involve errors beyond the reach of requirement rewriting---likely reflecting fundamental generation limitations for certain visualization types.

\begin{table*}[!htb]
\centering
\small
\setlength{\tabcolsep}{3pt}
\caption{CMV generation performance and annealing effectiveness across frontier LLMs. Per-component success rates (D/V/I) are shown as \textit{first-pass\,/\,post-spec-annealing\,/\,post-obj-annealing}. Req./Spec.\ columns show \textit{before/after} annealing success rates; Obj.\ is the final node success rate after object-level regeneration. Annealing columns show errors introduced (Int.), resolved (Res.), and resolution rate (RR.\%).}
\begin{tabular}{l ccccccc cc cc cc c r}
\toprule
\multirow{2}{*}{\textbf{Model}} & \multicolumn{7}{c}{\textbf{Generation Success Rate (\%)}} & \multicolumn{2}{c}{\textbf{Req. Annealing}} & \multicolumn{2}{c}{\textbf{Spec. Annealing}} & \multicolumn{2}{c}{\textbf{Obj. Annealing}} & \multirow{2}{*}{\textbf{RR.\%}} & \multirow{2}{*}{\textbf{Tokens}} \\
\cmidrule(lr){2-8} \cmidrule(lr){9-10} \cmidrule(lr){11-12} \cmidrule(lr){13-14}
& E2E & D & V & I & Req. & Spec. & Obj. & Res./Int. & RR.\% & Res./Int. & RR.\% & Res./Int. & RR.\% &  & \\
\midrule
Claude Sonnet 4.6  & \textbf{75.0} & 78.1/96.9/98.4 & \textbf{96.2}/\textbf{96.2}/\textbf{97.5} & \textbf{70.4}/\textbf{77.8}/\textbf{81.5} & 45.2/\textbf{100} & \textbf{85.4}/\textbf{93.6} & \textbf{95.3} & 40/40 & \textbf{100} & 13/22 & \textbf{59.1} & 2/10 & \textbf{20.0} & \textbf{76.4} & 280K \\
Gemini 3.1 Pro     & 50.0 & \textbf{83.9}/\textbf{100}/\textbf{100} & 78.7/78.7/78.7 & 46.2/50.0/50.0 & \textbf{57.6}/\textbf{100} & 74.8/81.8 & 81.8 & 28/28 & \textbf{100} & 10/23 & 43.5 & 0/13 & 0 & 59.4 & \textbf{133K} \\
GPT-5.4            & 41.7 & 49.2/83.1/84.6 & 84.1/84.1/85.5 & 63.3/66.7/70.0 & 44.8/\textbf{100} & 66.5/80.5 & 82.3 & 48/48 & \textbf{100} & 23/44 & 52.3 & 3/23 & 13.0 & 64.3 & 363K \\
Qwen-Plus          & 33.3 & 66.7/91.2/93.0 & 81.4/83.1/86.4 & 31.2/37.5/50.0 & 38.7/98.4 & 68.9/81.1 & 84.8 & 37/38 & 97.4 & 16/32 & 50.0 & 5/17 & 29.4 & 66.7 & 253K \\
DeepSeek-V3.2      & 25.0 & 50.0/75.9/81.0 & 72.4/74.1/79.3 & 35.0/35.0/45.0 & 42.2/96.9 & 57.4/69.1 & 75.0 & 35/37 & 94.6 & 16/43 & 37.2 & 8/25 & 32.0 & 56.2 & 301K \\
\midrule
Claude Code          & 8.33 & 100 & 100 & 48.0 & - & - & - & - & - & - & - & - & - & - & 783K \\
\bottomrule
\end{tabular}
\label{tab:benchmark_results}
\end{table*}

\subsection{Baseline Comparison}
\label{subsec:baseline}

To assess the advantage of structured generation over direct LLM-based coding, we compare \system{} against Claude Code, an agentic coding assistant with tool-calling and error-checking capabilities. Both systems use Claude Sonnet 4.6 as the foundation model, ensuring that performance differences reflect the generation framework rather than model capability. Both systems receive the same inputs: the original CSV tables, dataset descriptions, and analytical goals (reused from the benchmark). Claude Code additionally receives a prompt explaining what a CMV is and what the expected output format should be (a multi-view dashboard with data transformations, visualizations, and cross-view interactions), but is given no constraints on requirements or technical approach, allowing it to leverage its own harness engineering freely. For all interactive prompts, we accept default options and let the agent run until it terminates. To evaluate the output, we manually decompose Claude Code's generated dashboards into D, V, and I components following the same criteria used in the benchmark. A D component succeeds if its data transformation code executes without errors; a V component succeeds if its corresponding view renders correctly; an I component succeeds if its intended cross-view coordination logic (identified from the generated code) is actually realized in the rendered dashboard with correct behavior. E2E success requires all components to pass. Results are reported in the last row of \autoref{tab:benchmark_results}.

Claude Code achieves 100\% success on both D and V components, benefiting from its agentic retry loop that iteratively fixes code until execution succeeds. However, its I success rate drops to 48.0\% and E2E success to only 8.33\%---interactions are frequently intended in the code but fail to function correctly at runtime, revealing that interaction coordination---which requires cross-view field-level reasoning---cannot be reliably solved through generic code repair alone. In contrast, \system{} (Claude Sonnet 4.6) achieves 75.0\% E2E with 81.5\% I success post-obj-annealing, demonstrating that structured decomposition and layered validation provide substantially more effective guidance for cross-component consistency. Notably, Claude Code consumes an average of 783K tokens per task---over 2.8$\times$ more than \system{} (280K)---yet achieves far lower E2E success, confirming that brute-force agentic iteration is both less effective and less efficient than structured generation with semantic annealing.

\subsection{User Study}
\label{subsec:user_study}

The benchmark and baseline comparison evaluate generation correctness in a fully automated setting. The user study complements these by validating the key design decisions derived from our preliminary study (\autoref{sec:preliminary}): whether the D-V-I decomposition provides a comprehensible mental model for CMV construction (\textbf{DC1}), whether the progressive refinement workflow enables effective user control (\textbf{DC2, DC4}), and whether layered validation feedback supports efficient error resolution (\textbf{DC3}). Since no existing system offers a comparable structured CMV generation workflow with editable intermediate representations, we conduct a single-system evaluation rather than a comparative study. The study was approved by the university's institutional review board. Each session lasted 40--50 minutes.

\subsubsection{Participants}
We recruited 12 participants (7 male, 5 female) via social media and email, compensated at approximately \$16/hour. Participants included 8 PhD students and 4 master's students with 1--6 years of experience in data analysis, from diverse research domains including culture, sports, media studies, databases, computer vision, and machine learning. Rather than restricting recruitment to visualization specialists---who are scarce---we targeted data-intensive researchers who regularly perform exploratory data analysis and LLM-assisted coding (e.g., vibe coding, EDA) on their domain data, reflecting our intended user base of data analysis practitioners. All participants had prior experience with data analysis tools (e.g., Python/pandas, Matplotlib, Tableau) and reported daily use of LLM-based coding assistants (e.g., GitHub Copilot, Claude Code, Cursor), but varied in their familiarity with Vega-Lite and multi-view dashboard development.

\subsubsection{Procedure}

Each session consisted of three phases. In the \textit{introduction phase} ($\sim$10 minutes), we demonstrated the system's interface, explained the D-V-I decomposition, and walked through a sample CMV generation. In the \textit{exploration phase} ($\sim$15--25 minutes), each participant freely chose a dataset from the benchmark collection (e.g., earthquake, e-commerce, football) and provided an analytical goal---either self-formulated or auto-generated by the system. The system generated an initial CMV, after which participants iteratively refined the output: inspecting rendered views, editing requirement descriptions, triggering regeneration for individual nodes or failed sub-graphs, and testing cross-view interactions, until they were satisfied with the result or decided no further improvement was achievable. In the \textit{evaluation phase} ($\sim$15 minutes), participants completed a questionnaire and a semi-structured interview.

Participants rated six usability dimensions on a 7-point Likert scale (1~=~strongly disagree, 7~=~strongly agree), adapted from prior LLM-assisted visualization construction studies~\cite{deng2023dashbot,wu2021multivision} to cover the key facets of structured CMV generation: comprehensibility of intermediate representations, learnability, user control over generation, iteration efficiency, output correctness, and overall experience.
\begin{itemize}[leftmargin=*,nosep]
    \item \textbf{Transparency}: I can understand the D-V-I decomposition and the system's execution feedback.
    \item \textbf{Ease of use}: The interactive system is easy to learn and use.
    \item \textbf{Controllability}: I can effectively steer generation results.
    \item \textbf{Efficiency}: The system provides satisfactory feedback efficiently.
    \item \textbf{Output quality}: The generated CMVs are satisfactory.
    \item \textbf{Overall satisfaction}: The overall usage experience is satisfactory.
\end{itemize}
A semi-structured interview followed, probing participants' perceptions of the D-V-I decomposition, the requirement editing experience, the usefulness of annealing feedback, and suggestions for improvement. Three authors independently coded the interview transcripts and consolidated themes through discussion, following thematic analysis~\cite{braun2006using}.

\subsubsection{Results}

\begin{figure}[tb]
\centering
\includegraphics[width=\columnwidth]{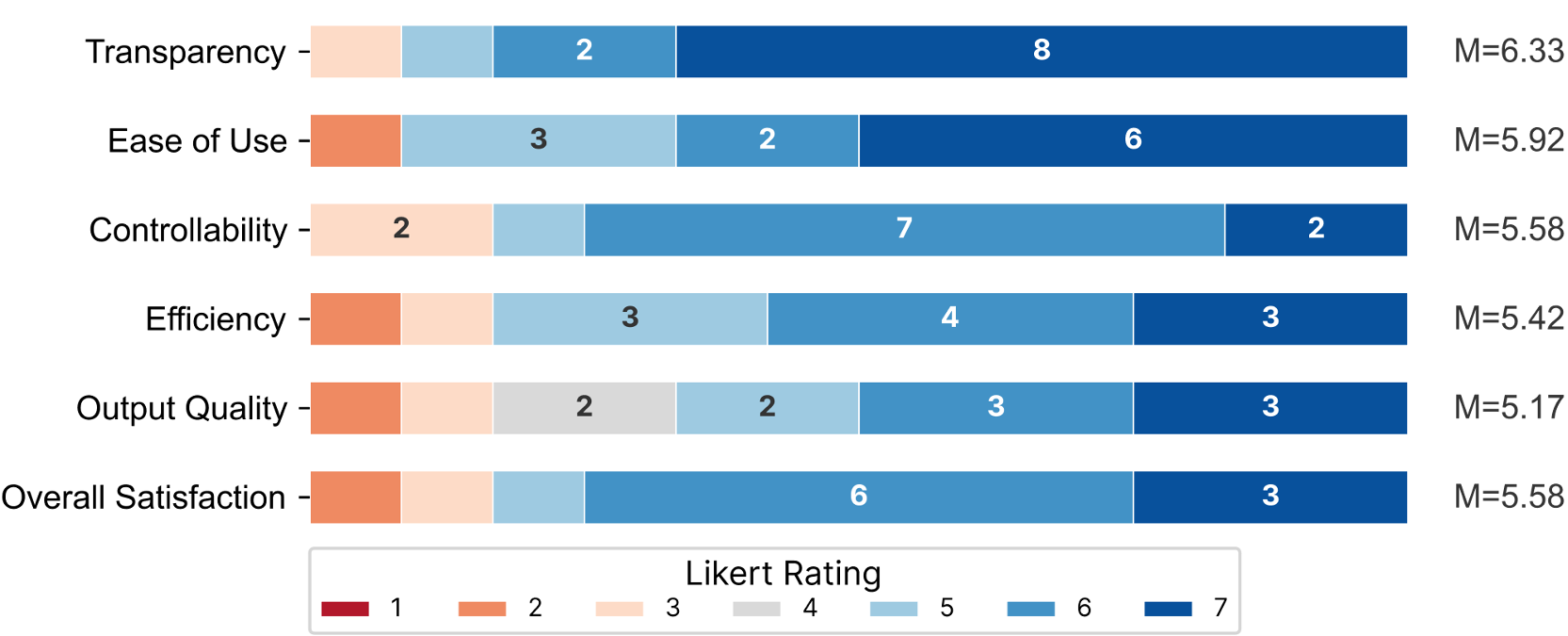}
\caption{User study Likert ratings (7-point scale, $N\!=\!12$). Each bar shows the distribution of responses; mean scores are annotated on the right.}
\label{fig:likert}
\end{figure}

\autoref{fig:likert} summarizes the Likert ratings. All six dimensions received mean scores above 5 on the 7-point scale, and 68\% of all individual ratings were 6 or 7. \textbf{Transparency} received the highest score ($M\!=\!6.33$, $SD\!=\!1.23$), with 10 of 12 participants rating it 6 or 7. \textbf{Ease of use} ($M\!=\!5.92$, $SD\!=\!1.51$) and \textbf{Controllability} ($M\!=\!5.58$, $SD\!=\!1.31$) also scored well. \textbf{Output quality} received the lowest yet still favorable score ($M\!=\!5.17$, $SD\!=\!1.64$). Overall satisfaction averaged $5.58$ ($SD\!=\!1.56$).

All 12 participants successfully constructed a usable CMV within 1--4 requirement edits and 1--5 regeneration cycles (11--26 minutes of generation time), confirming that the progressive refinement workflow enables efficient CMV prototyping with minimal iteration. Semi-structured interviews further validated the design decisions and revealed directions for future improvement.

\textbf{DC1: The D-V-I decomposition provides an effective mental model.}
All 12 participants found the decomposition logical and consistent with how they approach data analysis. P6 noted that it is ``\textit{especially helpful when the analytical goal is unclear or you don't know where to start---it helps break down the task.}'' P12 found the workflow ``\textit{intuitive and aligned with task decomposition thinking.}'' Participants generally prioritized inspecting V and I outputs first, turning to D nodes and requirement text only when results were unsatisfactory (P2, P9, P11).

\textbf{DC2\,\&\,DC4: Requirement editing enables effective steering.}
Participants successfully steered generation through requirement editing---P1 used bold text emphasis to direct the model's attention, and P2 found that both detailed and vague edits produced correct results. P7 resolved a cross-table field-format mismatch (``\textit{2024Q1}'' vs.\ ``\textit{2024-Q1}'') by tracing the annealing error from an I node back to the D requirement and editing the transformation description. P11 found feeding error messages back to the model ``\textit{reasonable and natural.}''

\textbf{DC3: Layered validation feedback supports error resolution.}
Annealing feedback was instrumental when participants encountered failures. Rather than opaque error messages, the layered checks (topology, field existence, semantic category) provided actionable diagnostics traceable through the dependency graph. This enabled even participants unfamiliar with Vega-Lite to diagnose and resolve cross-component issues through requirement editing alone.

Participants also identified directions for future improvement, including condensed requirement representations, richer interaction types, and dynamic node management; we discuss these in \autoref{sec:discussion}.

\section{Discussion}
\label{sec:discussion}

Our evaluation combines automated benchmarks with practitioner feedback, revealing both the strengths and boundaries of structured LLM-guided CMV generation. Below we distill cross-cutting lessons and outline directions for future work.

\subsection{Lessons Learned}

Five cross-cutting lessons emerge from the evaluation section.

\textbf{Structural correctness provides a foundation for domain-driven quality improvement.}
\system{} establishes that LLMs can reliably produce structurally correct, renderable CMVs with functioning cross-view interactions---a prerequisite that prior approaches could not consistently achieve. With this foundation, the next lever for improving CMV quality lies in upstream decisions---which transformations to apply, which views to create, how to coordinate them---that shape analytical value. These decisions hinge on domain understanding and analytical planning, and our user study confirms this: practitioners surfaced needs (richer interaction patterns, better requirement affordances) that reflect domain-level rather than structural concerns. Injecting domain knowledge---through retrieval-augmented generation or external knowledge sources---could inform these decisions, building on the structural guarantees that \system{} already provides.

\textbf{Feedback specificity matters more than retry budget.}
The benchmark and baseline comparison reveal a clear pattern: error correction effectiveness depends on the \textit{granularity of feedback}, not the number of retries. Requirement-level annealing resolves 94.6--100\% of topology errors with targeted structural checks, while specification-level annealing catches field-existence and semantic-category violations invisible to execution-based feedback. The baseline sharpens this contrast---Claude Code's agentic retry loop achieves 100\% D and V success through execution feedback alone, yet only 48\% I success, because runtime errors do not surface cross-view coupling violations. This suggests that structured generation frameworks should invest in domain-specific validation rather than longer retry loops.

\textbf{Interaction coordination is the primary bottleneck---both technically and experientially.}
I-Spec yields the lowest success rates across models (35.0--77.8\% post-spec-annealing), and I failures dominate E2E failures. The user study corroborates this: interaction issues were the most common frustration, and richer coordination patterns were the top improvement request (\autoref{subsec:user_study}). The convergence of automated and qualitative evidence points to multi-hop cross-view reasoning as a fundamental capability gap in current LLMs, suggesting that interaction generation may benefit from specialized strategies such as template-based coordination or interaction-specific fine-tuning.

\textbf{Structured decomposition aids both generation and human understanding.}
The D-V-I decomposition serves a dual purpose: it constrains LLM generation into manageable units with explicit contracts, and---as the user study confirms (\autoref{subsec:user_study})---provides practitioners with a mental model that aligns with their analytical thinking. This suggests that intermediate abstractions designed for machine generation can simultaneously serve as effective human-facing representations, a principle worth exploring in other structured generation domains.

\textbf{Rigid topology trades flexibility for verifiability.}
The framework constrains each D node to transform source tables into a single intermediate table, with no D$\rightarrow$D edges. Exploratory data analysis often chains transformations---producing multiple intermediate tables referencing each other---and the baseline confirms unconstrained agentic generation produces such patterns. However, this flexibility is also the primary source of errors: without explicit contracts at each intermediate step, field-level inconsistencies propagate silently. The core challenge is designing a more flexible \textit{orchestrator} that manages richer dependency topologies (e.g., D$\rightarrow$D chains), maintains context across intermediate steps, and validates structural, field-level, visual, and interaction correctness---balancing expressiveness with verifiability that makes structured generation reliable.

\subsection{Limitations and Future Work}

Our user study and benchmark analysis together surface several limitations that point to future research directions.

\textbf{Expressiveness of the output.}
\system{} targets Vega-Lite, limiting generated views to standard chart types. Custom visual designs---such as ForVizor's pitch overlay---cannot be produced. Extending the framework to more expressive grammars (e.g., D3) or coordination architectures such as Mosaic~\cite{heer2024mosaic} that support richer cross-view interactions through a shared database layer is a promising direction.

\textbf{Node management and iterative refinement.}
The current system generates a fixed set of D/V/I nodes from the initial goal. As user study participants noted (\autoref{subsec:user_study}), the inability to add or remove nodes after generation limits iterative exploration. Supporting dynamic node management and providing a topology overview would better align with the incremental nature of exploratory analysis.

\textbf{Scalability and evaluation breadth.}
Our benchmark contains 12 tasks, but each task produces 8--18 D/V/I nodes, yielding 130--170 individual generation-and-validation instances per model run---providing substantial per-component coverage despite the moderate task count. Nevertheless, the benchmark covers CMVs with up to approximately 6 views; scaling to larger dashboards may stress the incremental interaction generation strategy and increase cascading failures. Additionally, each model--task combination was evaluated in a single run; multiple runs would better quantify generation stability.

\section{Conclusion}

We presented \system{}, a framework that guides LLMs through CMV generation via \textit{query-centric CMV modeling}, \textit{progressive nucleation}, and \textit{semantic annealing}, achieving up to 75\% end-to-end success and substantially outperforming agentic code generation (75\% vs.\ 8.3\% E2E). Two key insights emerge: domain-specific validation matters more than retry budget---targeted structural and semantic checks resolve errors that unconstrained agentic iteration cannot; and structured decomposition simultaneously enables machine verifiability and human comprehension, though its rigid topology must evolve toward more adaptive orchestration. These findings suggest that layered intermediate abstractions can guide LLMs through complex multi-component generation tasks---a principle we believe extends beyond CMVs to other structured artifacts in visual analytics and software engineering.


\bibliographystyle{abbrv-doi-hyperref}
\bibliography{main}

\end{document}